\pdfoutput=1

\documentclass[11pt]{article}

\usepackage{coling}

\usepackage{times}
\usepackage{latexsym}

\usepackage[T1]{fontenc}

\usepackage[utf8]{inputenc}

\usepackage{microtype}

\usepackage{inconsolata}

\usepackage{graphicx}
\usepackage{caption}
\usepackage{subcaption}

\usepackage{pdflscape}
\usepackage{fancyhdr}
\usepackage{booktabs}
\usepackage{array}
\usepackage{setspace}

\newcolumntype{P}[1]{>{\setstretch{0.85}\raggedright\arraybackslash}p{#1}}

\usepackage{everypage-1x}
\usepackage{lipsum}
\usepackage{geometry}

\usepackage{float}            
\usepackage{tcolorbox} 
\tcbuselibrary{breakable}

\usepackage{enumitem}

\usepackage{amsmath}
\usepackage{amssymb}
\usepackage{natbib}

\usepackage{hyperref}

\usepackage{comment}

\usepackage{xcolor}

\title{Neuro-Conceptual Artificial Intelligence: Integrating OPM with Deep Learning to Enhance Question Answering Quality}

\author{Xin Kang$^{1}$, Veronika Shteingardt$^{2}$, Yuhan Wang$^{1}$, and Dov Dori$^{2}$ \\
  $^{1}$Tokushima University, Tokushima, Japan \\
  $^{2}$Technion – Israel Institute of Technology, Haifa, Israel \\
  \texttt{kang-xin@is.tokushima-u.ac.jp, veronika-s@campus.technion.ac.il,} \\
  \texttt{c612494013@tokushima-u.ac.jp, dori@technion.ac.il}}

\begin{document}
\maketitle
\begin{abstract}
Knowledge representation and reasoning are critical challenges in Artificial Intelligence (AI), particularly in integrating neural and symbolic approaches to achieve explainable and transparent AI systems. Traditional knowledge representation methods often fall short of capturing complex processes and state changes. We introduce \textbf{Neuro-Conceptual Artificial Intelligence (NCAI)}, a specialization of the neuro-symbolic AI approach that integrates conceptual modeling using Object-Process Methodology (OPM) ISO 19450:2024 with deep learning to enhance question-answering (QA) quality. By converting natural language text into OPM models using in-context learning, NCAI leverages the expressive power of OPM to represent complex OPM elements—processes, objects, and states—beyond what traditional triplet-based knowledge graphs can easily capture. This rich structured knowledge representation improves reasoning transparency and answer accuracy in an OPM-QA system. We further propose transparency evaluation metrics to quantitatively measure how faithfully the predicted reasoning aligns with OPM-based conceptual logic. Our experiments demonstrate that NCAI outperforms traditional methods, highlighting its potential for advancing neuro-symbolic AI by providing rich knowledge representations, measurable transparency, and improved reasoning.
\end{abstract}

\section{Introduction}
\label{sec:introduction}

Integrating neural and symbolic approaches in AI seeks to combine the learning capabilities of neural networks with the interpretability of symbolic reasoning \citep{Besold2017NeuralSymbolic, Garcez2023NeuroSymbolic}. However, traditional knowledge representations, such as triplet-based knowledge graphs, are limited in capturing complex processes, state changes, and hierarchical relationships inherent in dynamic systems \citep{Wang2017KnowledgeGraphEmbedding, heinzerling-inui-2021-language, shi-etal-2021-transfernet}. Additionally, neural networks are often viewed as a black box due to their opaque decision-making processes, which poses significant challenges in domains requiring transparent reasoning, such as healthcare and finance \citep{Lipton2018Mythos, Rudin2019StopExplaining, DoshiVelez2017RigorousScience, Tjoa2020SurveyXAI}.

\begin{figure}[t!]
    \centering
    \includegraphics[width=\linewidth]{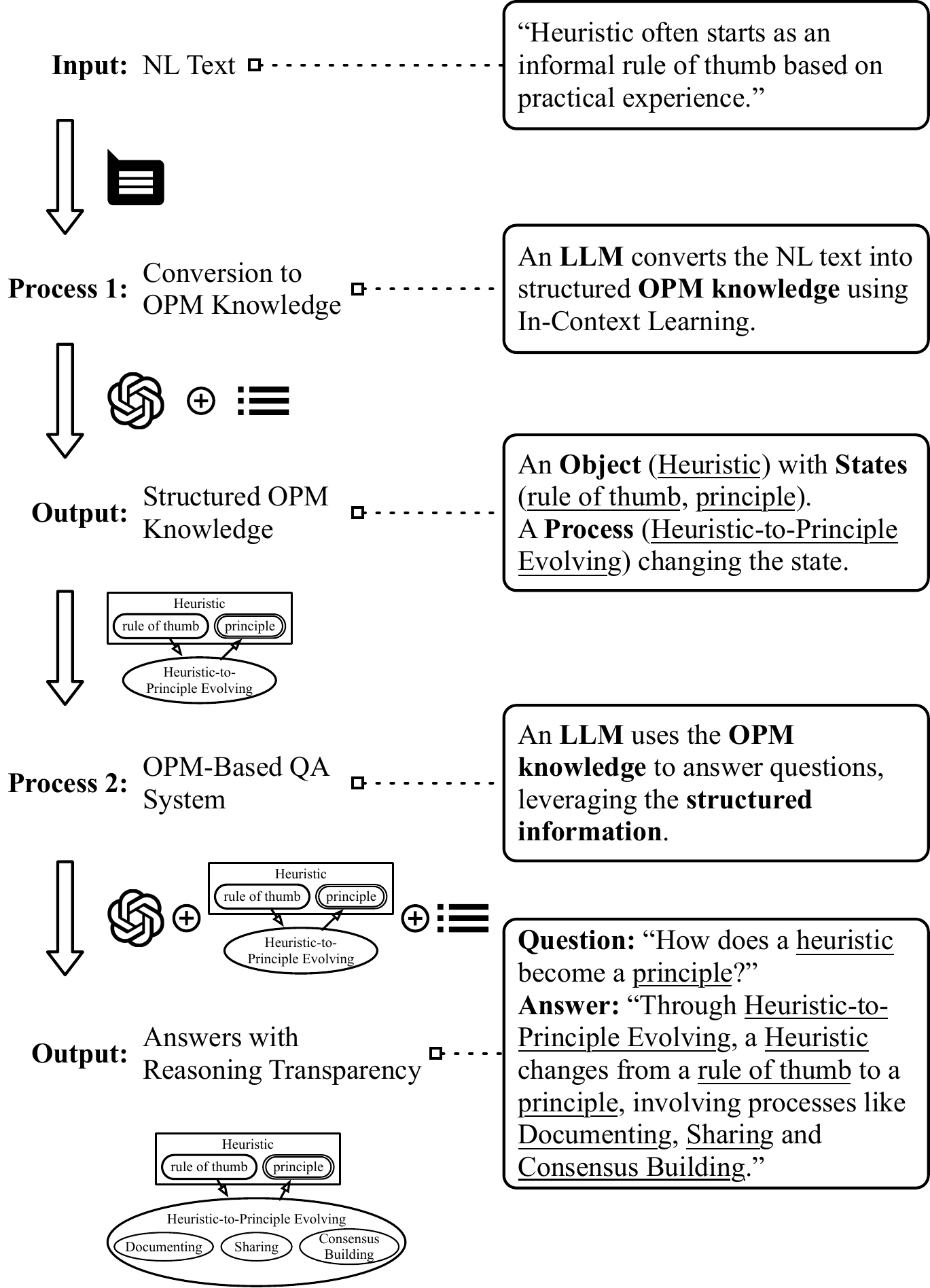}
    \caption{Overview of the NCAI framework, illustrating how the LLM converts natural language text into structured OPM knowledge and uses it in OPM-QA for transparent reasoning. Starting from the text ``Heuristic often starts as an informal rule of thumb \dots'', the model generates an OPM model and answers questions by referencing processes like \textit{Heuristic-to-Principle Evolving}.}
    \label{fig:overview}
\end{figure}

Recent advancements have focused on enhancing AI reasoning capabilities by integrating language models with external knowledge sources. For example, combining language models with knowledge graphs has been applied to improve question answering systems \citep{yasunaga-etal-2021-qa, oguz-etal-2022-unik, shi-etal-2021-transfernet, zhang-etal-2023-cikqa}. Despite these efforts, fully capturing dynamic behaviors and providing transparent reasoning paths remains a challenge.

To address these limitations, we introduce \textbf{Neuro-Conceptual Artificial Intelligence (NCAI)}, a specialization of the neuro-symbolic AI approach, in which the symbolic component is an Object-Process Methodology (OPM ISO 19450:2024) conceptual model. OPM is a conceptual modeling language and methodology that unifies the system's structural and behavioral aspects within a single model \citep{dori2002object, dori2016model}. It represents objects (things that exist) and processes (things that transform objects) in both graphical and textual modalities. By combining OPM with the large language model (LLM), NCAI enhances reasoning transparency and answer accuracy in QA tasks.

An overview of the NCAI framework is illustrated in Figure~\ref{fig:overview}. The framework begins by converting natural language text into structured OPM knowledge using in-context learning with an LLM. This structured knowledge is then used in an OPM-QA, which leverages the expressive power of OPM to represent complex processes and state changes that traditional triplet-based knowledge graphs cannot easily capture. By integrating conceptual modeling with deep learning, NCAI creates a pipeline that transforms unstructured text into a rich knowledge representation, enabling more effective AI reasoning and interpretability.

Our contributions in this work are threefold:

(1) We propose NCAI, which integrates OPM with deep learning to enhance reasoning transparency and answer accuracy.

(2) We develop OPM-QA that utilizes OPM knowledge to improve question-answering quality.

(3) We introduce transparency evaluation metrics to quantitatively assess how faithfully the predicted reasoning aligns with the conceptual logic defined by OPM, and we demonstrate the effectiveness of NCAI through experiments showing improved performance over traditional methods.

\begin{figure*}[t!]
    \centering
    \begin{subfigure}[b]{0.5\linewidth}
        \centering
        \includegraphics[width=\linewidth]{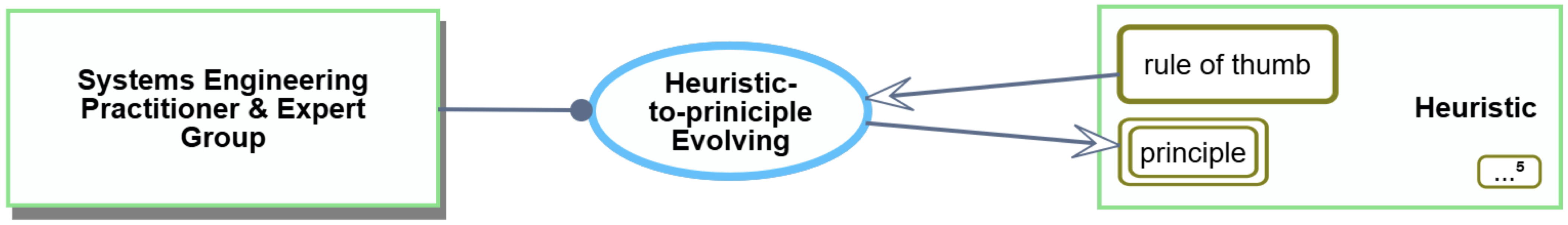}
        \caption{System Diagram (SD)}
        \label{fig:manual_opm_sd}
    \end{subfigure}

    \vspace{1em}

    \begin{subfigure}[b]{0.8\linewidth}
        \centering
        \includegraphics[width=\linewidth]{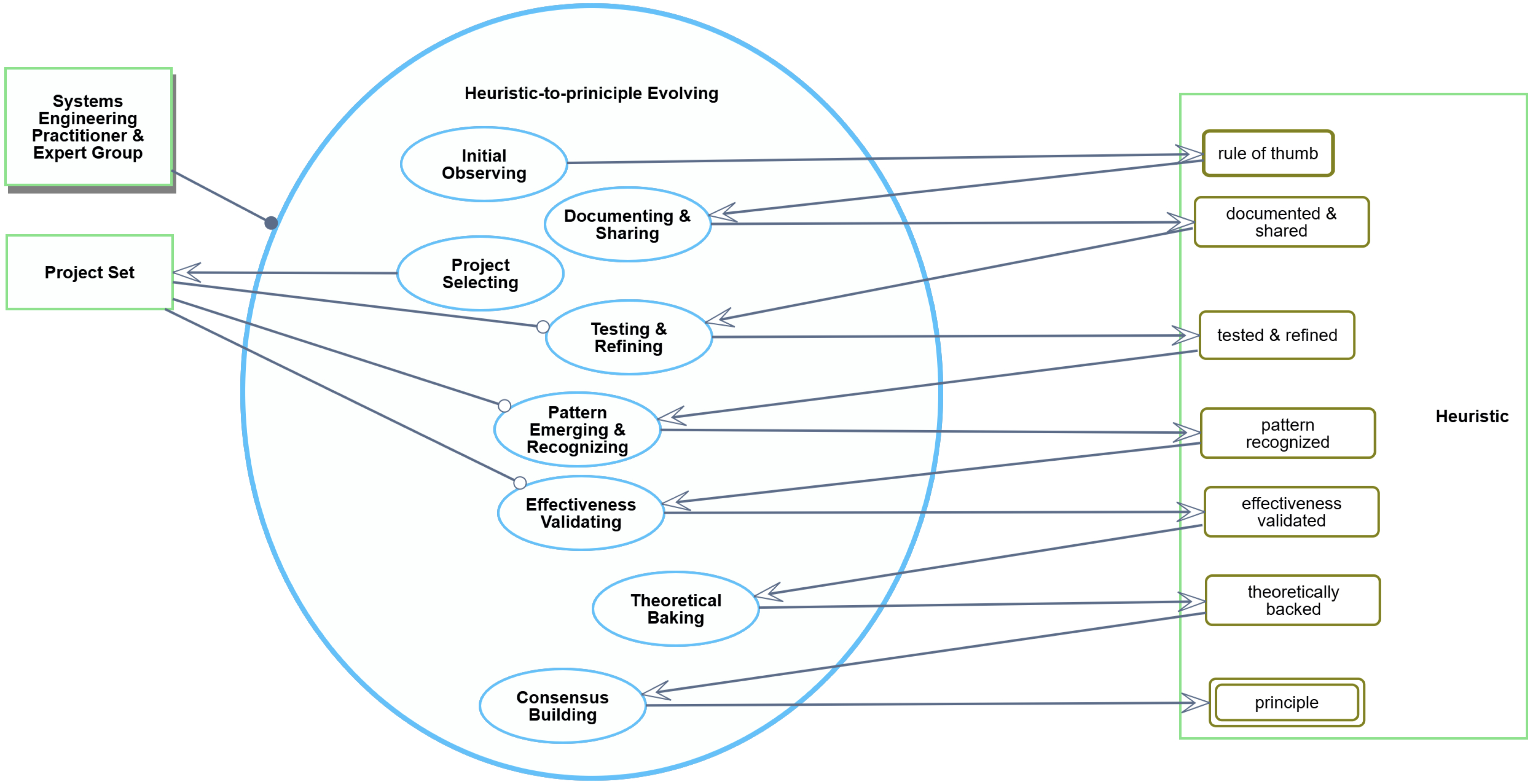}
        \caption{In-Zoomed Diagram (SD1)}
        \label{fig:manual_opm_sd1}
    \end{subfigure}

    \caption{Constructed OPDs illustrating the transformation of a \textit{Heuristic} from a \textit{rule of thumb} to a \textit{principle} through various OPM elements—processes, objects, and states—within the OPM framework.}
    \label{fig:manual_opm_model}
\end{figure*}

\section{Related Work}
\label{sec:related}

\paragraph{Neuro-Symbolic AI Approaches}

Neuro-symbolic AI integrates neural networks with symbolic reasoning to harness the strengths of both paradigms \citep{Besold2017NeuralSymbolic, Garcez2023NeuroSymbolic}. Challenges in achieving reasoning transparency and interpretability persist, with approaches such as symbolic knowledge distillation \citep{west-etal-2022-symbolic} and factual knowledge editing \citep{de-cao-etal-2021-editing} addressing these issues. Frameworks like TransferNet \citep{shi-etal-2021-transfernet} and interpretable reasoning models for dialogue generation \citep{yang-etal-2022-interpretable} aim to provide clear reasoning paths. In sentiment analysis and mental health, neuro-symbolic frameworks like TAM-SenticNet \citep{dou2024tam} and causal inference models \citep{ding2024causal, ding2024neuro} enhance explainability and logical inference. Specifically, in aspect-based sentiment analysis (ABSA), models such as the Multi-Agent Collaboration (MAC) \citep{kang-etal-2024-tmak} and approaches to improve AI transparency using generative agents \citep{kang2024transparency} demonstrate the potential of neuro-symbolic AI in providing transparent and rational sentiment analysis.

\paragraph{Interpretability and Transparency in Language Models}

Ensuring transparency and interpretability in AI decision-making is critical, particularly in complex systems \citep{Lipton2018Mythos, Rudin2019StopExplaining}. Various methods have been developed to enhance the interpretability of language models, including representation dissimilarity measures \citep{brown-etal-2023-understanding}, SHAP-based explanation techniques \citep{mosca-etal-2022-shap}, and prompt-based explainers like PromptExplainer \citep{feng-etal-2024-promptexplainer}. Evaluation benchmarks for interpretability \citep{wang-etal-2022-fine} and approaches to improve faithfulness and robustness \citep{el-zini-awad-2022-beyond, horovicz-goldshmidt-2024-tokenshap, zhao-etal-2024-tapera} further contribute to making language models more transparent. Despite these advancements, achieving full transparency remains challenging, especially in applications requiring a clear understanding of the reasoning process.

\paragraph{Language Models and Knowledge Graphs for Question Answering}

Integrating language models with knowledge graphs has been a significant focus to enhance QA capabilities. Approaches like QA-GNN \citep{yasunaga-etal-2021-qa}, DRLK \citep{zhang-etal-2022-drlk}, and UniK-QA \citep{oguz-etal-2022-unik} combine language models with graph neural networks and dynamic interactions to improve reasoning in QA tasks. Frameworks such as CIKQA \citep{zhang-etal-2023-cikqa} and Triple-R \citep{kanaani-etal-2024-triple} emphasize the integration of external knowledge sources for more accurate and interpretable reasoning. Additionally, methods like TaPERA \citep{zhao-etal-2024-tapera} enhance faithfulness and interpretability in long-form table QA through content planning and execution-based reasoning. These integrations, while improving performance, often involve complex architectures and still face challenges in achieving complete reasoning transparency.

\section{NCAI Framework}
\label{sec:ncai_framework}

\subsection{Object-Process Methodology for NCAI}

OPM unifies objects and processes within a single model, representing structural and behavioral aspects in both graphical and textual forms \citep{dori2002object, dori2016model}. OPM's bimodal property provides Object-Process Diagram (OPD) and Object-Process Language (OPL), enhancing understanding and reasoning transparency.

To illustrate OPM's capabilities, we use a running example based on natural language text describing the evolution of a \textit{Heuristic} from a \textit{rule of thumb} to a \textit{principle}. This text serves as the input to the NCAI framework, as shown in Figure~\ref{fig:overview}, and is provided in Appendix~\ref{sec:nlt}.

Using this text, we constructed OPDs representing the processes and state changes of a \textit{Heuristic}. The diagrams can be created and visualized using the OPCloud software \citep{dori2018object, kohen2021designing}.

Figure~\ref{fig:manual_opm_model} presents the constructed OPDs. The System Diagram (SD) in Figure~\ref{fig:manual_opm_sd} captures the overall transformation of object \textit{Heuristic} from state \textit{rule of thumb} to state \textit{principle} through the process \textit{Heuristic-to-Principle Evolving}. The In-Zoomed Diagram (SD1) in Figure~\ref{fig:manual_opm_sd1} provides a detailed view of the subprocesses involved, such as \textit{Documenting \& Sharing}, \textit{Testing \& Refining}, \textit{Pattern Emerging \& Recognizing}, \textit{Effectiveness Validating}, \textit{Theoretical Backing}, and \textit{Consensus Building}.

The corresponding OPL for the System Diagram (SD) and the In-Zoomed Diagram (SD1) are presented in Appendix~\ref{sec:ocom} . These OPLs provide a textual representation that details the processes and state changes of the evolution of a heuristic from a rule of thumb to a principle.

OPM's bimodal property, combining graphical OPD and textual OPL, facilitates a comprehensive representation of complex processes and state changes. The in-zooming mechanism allows for hierarchical decomposition, where processes can be detailed further in subsequent diagrams, enhancing understanding of intricate systems.

\subsection{Converting Natural Language to OPM using In-Context Learning}
\label{sec:nl_to_opm}

We employ in-context learning to guide the LLM in converting natural language text into OPM models. The process involves providing the LLM with a carefully crafted prompt that includes OPM syntax, semantics, and examples. The prompt details can be found in \citep{dori2025converting}.

Let $T_{\text{NL}}$ be the natural language text (Appendix~\ref{sec:nlt}) and $P_{\text{OPM}}$ the prompt containing OPM instructions and examples. The input to the LLM is:
\begin{equation}
I = P_{\text{OPM}} \circ T_{\text{NL}},
\end{equation}
where $\circ$ denotes concatenation. The LLM generates the OPL representation:
\begin{equation}
T_{\text{OPL}} = \text{LLM}(I).
\end{equation}

This process leverages the LLM's ability to generate structured textual OPL representations from unstructured text, utilizing in-context learning to guide the model's output toward the desired OPL format. The OPL generated by the LLM is presented in Appendix~\ref{sec:ogl}.

While the preliminary results are encouraging, designing prompts that yield accurate and syntactically correct OPM models from free-form text introduces several challenges. These include pinpointing the primary process, focusing on essential conceptual elements, and clarifying ambiguous relationships in the natural language source. To address these issues, we iteratively refine prompts, adjust instructions, and incorporate carefully chosen examples. Through this iterative approach, the LLM learns to better navigate textual ambiguities and produce more coherent OPM models, thus reducing the need for extensive manual refinement and enabling more reliable neuro-symbolic reasoning pipelines.

\subsection{OPM Knowledge-Based Question-Answering System}
\label{sec:opm_qa_system}

We developed OPM-QA, an OPM knowledge-based Question-Answering system, that integrates OPM knowledge with the LLM to enhance answer accuracy and reasoning transparency. This system is a core component of NCAI, leveraging the structured knowledge representation of OPM to improve the reasoning capabilities of the LLM.

OPM-QA employs in-context learning by providing the LLM with OPL as the OPM knowledge, a set of example question-answer pairs, and the test questions as context. The knowledge $K_{\text{OPL}}$ is derived from the constructed OPM (see Appendix~\ref{sec:ocom}) and provides a structured and formalized representation. This structured knowledge allows the LLM to reason more effectively when generating answers.

For each test question $q_i$ in the set of test questions $Q_{\text{test}}$, the input to the LLM is formulated as:

\begin{equation}
I_i = K_{\text{OPL}} \circ E_{\text{QA}} \circ q_i,
\end{equation}
where $E_{\text{QA}}$ is the set of example question-answer pairs, and $\circ$ denotes concatenation. The LLM processes this input and generates an answer:

\begin{equation}
a_i = \text{LLM}(I_i).
\end{equation}

To assess the impact of using structured OPM knowledge on the QA performance, we compare the OPM-QA with a baseline QA system using natural language knowledge (NL-QA). In NL-QA, we replace $K_{\text{OPL}}$ with the natural language knowledge $K_{\text{NL}}$, which corresponds to the text provided in Appendix~\ref{sec:nlt}. This allows us to compare the effectiveness of the structured OPM knowledge against unstructured natural language knowledge in the QA task.

\section{Experiments}
\label{sec:experiments}

\subsection{Experiment Setup}

The purpose of our experiment is to evaluate the effectiveness of the NCAI framework in performing multi-hop reasoning tasks and enhancing reasoning transparency. We aim to compare the performance of OPM-QA with the baseline NL-QA.

\paragraph{Data:} We manually developed a dataset of 50 multi-hop reasoning question-answer pairs, following the FanOutQA benchmark \citep{zhu-etal-2024-fanoutqa}. These questions are based on the knowledge of the process that transforms informal rules of thumb into well-established principles. The questions require the model to integrate information from multiple statements to arrive at an answer, testing both answer accuracy and reasoning transparency. Examples of the QA pairs are provided in Appendix~\ref{appendix:qa_results}, Table~\ref{tab:qa_table}.

\paragraph{Knowledge Sources:} The OPL knowledge \( K_{\text{OPL}} \) is the OPL generated from the constructed OPM model in Appendix~\ref{sec:ocom}. The natural language knowledge \( K_{\text{NL}} \) is the text provided in Appendix~\ref{sec:nlt}. The QA systems use either \( K_{\text{OPL}} \) or \( K_{\text{NL}} \), along with 5 example QA pairs \( E_{\text{QA}} \) as context, to answer the 50 test questions \( Q_{\text{test}} \).

\paragraph{QA Systems:} The QA systems employ in-context learning by providing the LLM with the respective knowledge source, a set of example QA pairs, and the test questions. The LLM used for both systems is \texttt{GPT-4o} (version \texttt{o1-preview-2024-09-12}), with parameters set to $\mathit{temperature}=0$ and $\mathit{top\_p}=1$ to ensure deterministic output. By using the same LLM and parameter settings, we ensure a fair comparison between OPM-QA and NL-QA. The prompt used in these QA systems is shown in Appendix~\ref{sec:qa_prompt}. It has been carefully designed to be general enough for QA tasks across various domains and yet instructive enough to guide the model to answer with explicit reference to the OPM elements—processes, objects, and states, thereby increasing reasoning transparency and answer accuracy.

\paragraph{Evaluation Metrics:} We evaluate system outputs using a combination of metrics that capture different aspects of answer quality and reasoning transparency. To assess how well the generated answers align with the ground truth in terms of content, we use Loose Accuracy and Strict Accuracy. Loose Accuracy measures the fraction of reference tokens that also appear in the predicted answer after lemmatization, removing stop words, and stripping punctuation, providing a relatively lenient measure of correctness. Strict Accuracy applies a non-linear weighting (with a parameter $k=1.5$) to penalize partial matches more severely, thus enforcing a stricter standard of correctness.

While Loose Accuracy and Strict Accuracy focus on token-level overlap, ROUGE-1 (R-1), ROUGE-2 (R-2), and ROUGE-L (R-L) \citep{lin-2004-rouge} quantify lexical overlap through $n$-gram and sequence-based comparisons, capturing syntactic similarity between the generated answer and the reference. BLEURT (BT) \citep{sellam-etal-2020-bleurt} complements these metrics by providing a more semantic-oriented evaluation, as it uses a learned model to judge the meaning and quality of the generated text. The GPT Judge Score (GPT) \citep{zhu-etal-2024-fanoutqa} further evaluates factual consistency and logical coherence, reflecting how well the answer maintains internal logical structure and correctness from a large language model’s perspective.

To address the need for a quantitative measure of reasoning transparency, we propose Transparency Precision ($\text{P}_\text{T}$), Transparency Recall ($\text{R}_\text{T}$), and Transparency F1 ($\text{F1}_\text{T}$). Let $\mathcal{E}_{p}$ be the set of OPM elements-processes, objects, and states-identified in the prediction, and $\mathcal{E}_{g}$ the set of OPM elements in the ground truth. Let $\mathcal{E}_{p \cap g}$ be the intersection of these sets, representing correctly matched OPM elements. We define:

\begin{align}
\text{P}_\text{T} &= \frac{|\mathcal{E}_{p \cap g}|}{|\mathcal{E}_{p}|}, \\[6pt]
\text{R}_\text{T} &= \frac{|\mathcal{E}_{p \cap g}|}{|\mathcal{E}_{g}|}, \\[6pt]
\text{F1}_\text{T} &= \frac{2 \cdot \text{P}_\text{T} \cdot \text{R}_\text{T}}{\text{P}_\text{T} + \text{R}_\text{T}}.
\end{align}

Here, $\text{P}_\text{T}$ measures how accurately the predicted reasoning structure identifies the correct OPM elements, $\text{R}_\text{T}$ gauges how completely it recovers them, and $\text{F1}_\text{T}$ balances both. Together, these transparency metrics provide a statistical measure of how faithfully the system’s reasoning aligns with the conceptual logic defined by OPM, offering a principled, quantitative response to calls for more objective assessments of reasoning transparency.

\subsection{Results}

\begin{table*}[t!]
\centering
\begin{tabular}{lccc}
\toprule
\textbf{Metric} & \textbf{OPM-QA} & \textbf{NL-QA} & \textbf{P-value} \\
\midrule
Loose Accuracy & \textbf{0.858 $\pm$ 0.162} & 0.638 $\pm$ 0.212 & $< 0.001$ \\
Strict Accuracy & \textbf{0.806 $\pm$ 0.213} & 0.530 $\pm$ 0.252 & $< 0.001$ \\
ROUGE-1 & \textbf{0.772 $\pm$ 0.159} & 0.558 $\pm$ 0.195 & $< 0.001$ \\
ROUGE-2 & \textbf{0.607 $\pm$ 0.201} & 0.373 $\pm$ 0.198 & $< 0.001$ \\
ROUGE-L & \textbf{0.715 $\pm$ 0.155} & 0.504 $\pm$ 0.174 & $< 0.001$ \\
BLEURT & \textbf{0.596 $\pm$ 0.165} & 0.474 $\pm$ 0.111 & $< 0.001$ \\
GPT Judge Score & \textbf{0.920 $\pm$ 0.274} & 0.800 $\pm$ 0.404 & 0.086 \\
Transparency Precision & \textbf{0.917 $\pm$ 0.161} & 0.759 $\pm$ 0.417 & 0.015 \\
Transparency Recall & \textbf{0.953 $\pm$ 0.143} & 0.455 $\pm$ 0.329 & $< 0.001$ \\
Transparency F1 & \textbf{0.922 $\pm$ 0.136} & 0.546 $\pm$ 0.342 & $< 0.001$ \\
\bottomrule
\end{tabular}
\caption{Evaluation results comparing OPM-QA and NL-QA across correctness, lexical similarity, semantic quality, factual consistency, and transparency. P-values indicate that OPM-QA significantly outperforms NL-QA on all metrics with high statistical confidence, except for GPT Judge and Transparency Precision, where the differences are less significant.}
\label{tab:qa_results}
\end{table*}

Table~\ref{tab:qa_results} presents the results of our evaluation. For Loose Accuracy, OPM-QA achieves 0.858 $\pm$ 0.162, greatly exceeding NL-QA’s 0.638 $\pm$ 0.212. This indicates that OPM-QA captures a significantly larger fraction of reference tokens under a lenient matching criterion. The difference is statistically significant (P < 0.001). Strict Accuracy, which imposes a harsher penalty on partial matches, shows OPM-QA at 0.806 $\pm$ 0.213 compared to NL-QA's 0.530 $\pm$ 0.252. This improvement is also statistically significant (P < 0.001), demonstrating that OPM-QA provides answers that are both more complete and more precisely aligned with the ground truth.

Regarding syntactic overlap measures, OPM-QA significantly outperforms NL-QA in all ROUGE metrics. The ROUGE-1 score for OPM-QA is 0.772 $\pm$ 0.159 versus NL-QA's 0.558 $\pm$ 0.195, ROUGE-2 is 0.607 $\pm$ 0.201 compared to 0.373 $\pm$ 0.198, and ROUGE-L is 0.715 $\pm$ 0.155 compared to 0.504 $\pm$ 0.174. All these differences are highly statistically significant (P < 0.001). These results confirm that OPM-QA's generated answers exhibit considerably more lexical and subsequence-level similarity to the reference answers, adhering better to the structural and phrasing patterns of the ground truth.

In terms of semantic quality, the BLEURT score for OPM-QA is 0.596 $\pm$ 0.165, which surpasses NL-QA's 0.474 $\pm$ 0.111. This difference is statistically significant (P < 0.001). This suggests that OPM-QA not only matches lexically but also maintains closer semantic fidelity to the intended meanings of the ground truth answers.

Factual consistency and logical coherence are further evidenced by the GPT Judge Score of 0.920 $\pm$ 0.274 for OPM-QA compared to NL-QA's 0.800 $\pm$ 0.404. This difference is not statistically significant (P = 0.086), although it still indicates a notable improvement in maintaining factual and logical integrity within the answers.

Most notably, the transparency metrics reveal OPM-QA’s substantial advantage in conceptual alignment. OPM-QA achieves a Transparency Precision of 0.917 $\pm$ 0.161 and Transparency Recall of 0.953 $\pm$ 0.143, whereas NL-QA scores 0.759 $\pm$ 0.417 and 0.455 $\pm$ 0.329, respectively. The Precision difference is statistically significant (P = 0.015), while Recall remains highly significant (P < 0.001). Consequently, Transparency F1 for OPM-QA is 0.922 $\pm$ 0.136 compared to NL-QA's 0.546 $\pm$ 0.342, with a P-value of P < 0.001. This metric, which balances Transparency Precision and Transparency Recall, underscores the overall superior performance of OPM-QA in aligning with the ground truth both accurately and comprehensively.

Overall, the majority of these metrics demonstrate statistically significant improvements, affirming the superior performance of OPM-QA over NL-QA. Additionally, the enhancements in Transparency Precision metrics, despite being less statistically significant, further highlight OPM-QA’s effectiveness in achieving greater factual consistency and precision in answers. Detailed evaluation results for 10 representative QA examples and additional evaluation tables are provided in Appendix~\ref{appendix:qa_results}, including Tables~\ref{tab:qa_table}, \ref{tab:results_opl}, and \ref{tab:results_nl}, which further confirm these findings.

\subsection{Discussion}

The experimental results confirm that grounding the reasoning process in a conceptual model leads to both improved accuracy and clearer interpretability. Compared to its counterpart, the OPM-QA system consistently aligns its reasoning with the well-defined ontology provided by the OPM model. While the NL-QA system may occasionally produce correct or partially correct answers, it often does so without revealing the underlying conceptual structure. In contrast, OPM-QA not only identifies the correct OPM elements-processes, objects, and states-required to transform the heuristic from one state to another but also presents a reasoning chain that is faithful to the conceptual logic defined by OPM.

\begin{table*}[htbp]
\centering
\begin{tabular}{p{0.12\linewidth} p{0.8\linewidth}}
\toprule
\textbf{Question} & What processes change \textit{Heuristic} from \textit{documented \& shared} to \textit{theoretically backed}? \\
\midrule
\textbf{$\text{Answer}^{\text{GT}}$} & \textit{Heuristic} changes from \textit{documented \& shared} to \textit{theoretically backed} through \textit{\textcolor{blue}{Testing \& Refining}}, \textit{\textcolor{blue}{Pattern Emerging \& Recognizing}}, \textit{\textcolor{blue}{Effectiveness Validating}}, and \textit{\textcolor{blue}{Theoretical Backing}}. \\
\midrule
\textbf{$\text{Answer}^{\text{OPL}}$} & The processes that change \textit{Heuristic} from \textit{documented \& shared} to \textit{theoretically backed} are \textit{\textcolor{blue}{Testing \& Refining}}, \textit{\textcolor{blue}{Pattern Emerging \& Recognizing}}, \textit{\textcolor{blue}{Effectiveness Validating}}, and \textit{\textcolor{blue}{Theoretical Backing}}. \\
\midrule
\textbf{$\text{Answer}^{\text{NL}}$} & The processes that change \textit{Heuristic} from \textit{documented \& shared} to \textit{theoretically backed} are Testing \& Refinement, Pattern Recognition, Formal Studies, and \textit{\textcolor{blue}{Theoretical Backing}}. \\
\bottomrule
\end{tabular}
\caption{Comparison of ground truth answer $\text{Answer}^{\text{GT}}$ and answers from OPM-QA and NL-QA $\text{Answer}^{\text{OPL}}$ and $\text{Answer}^{\text{NL}}$ for a sample question, highlighting the matched processes in \textcolor{blue}{blue} corresponding to the reasoning transparency. While the ground truth and OPM-QA specify all relevant processes, NL-QA mentions only one correct process (\textit{\textcolor{blue}{Theoretical Backing}}). OPM-QA demonstrates a complete and conceptually aligned reasoning structure, whereas NL-QA’s reasoning chain remains incomplete.}
\label{tab:answer_comparison}
\end{table*}

Table~\ref{tab:answer_comparison} illustrates a representative case where the question focuses on the processes that guide the heuristic from a documented and shared state to a theoretically backed one. The ground truth answer specifies all of the required processes involved in this transformation. OPM-QA’s answer successfully enumerates each of these processes, maintaining exact alignment with the conceptual elements defined by the OPM model. In doing so, OPM-QA achieves high transparency metrics, as measured by the previously defined precision, recall, and F1 scores for transparency. Conversely, NL-QA identifies fewer correct conceptual elements, and in some cases introduces extraneous or irrelevant processes. This discrepancy highlights not merely a difference in correctness, but also a fundamental gap in the clarity and coherence of the reasoning steps offered by the two QA systems.

\begin{figure*}[t!]
    \centering
    \includegraphics[width=0.8\linewidth]{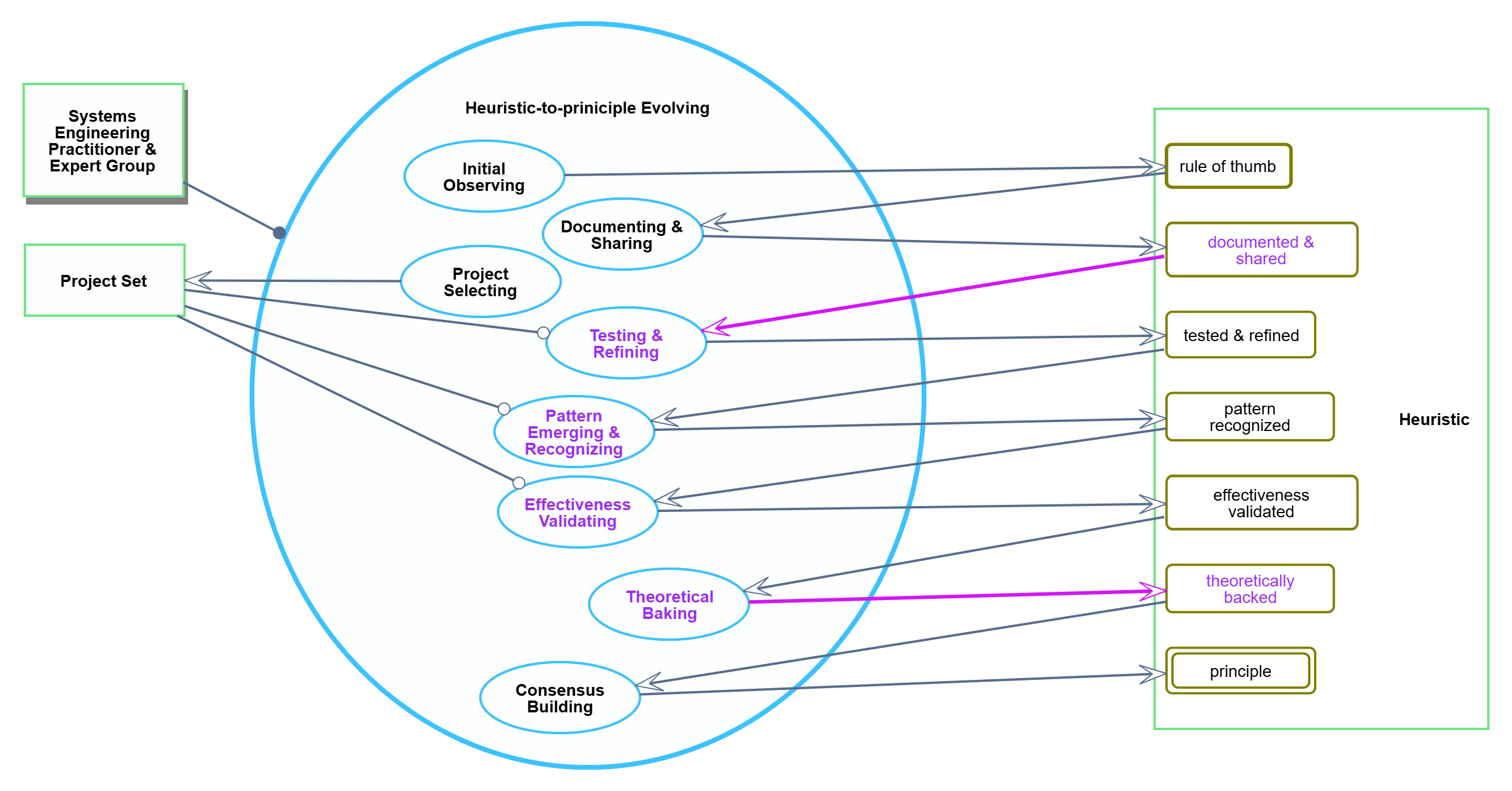}
    \caption{In-Zoomed Diagram (SD1) highlighting the specific processes in \textcolor{blue}{blue} involved in transforming \textit{Heuristic} from \textit{documented \& shared} to \textit{theoretically backed}. These highlighted processes match exactly those identified by OPM-QA in Table~\ref{tab:answer_comparison}, demonstrating a coherent and transparent reasoning path.}
    \label{fig:opg_example}
\end{figure*}

In addition to the tabular comparison, Figure~\ref{fig:opg_example} visually confirms that OPM-QA’s reasoning pathway closely follows the conceptual map provided by OPM. The figure displays an in-zoomed portion of the OPM model (SD1), where the processes critical to changing the heuristic’s state are clearly marked. Each process chosen by OPM-QA is found exactly where it should be according to the conceptual model. Observing these elements in the figure shows that OPM-QA’s improved transparency metrics correspond to verifiable reasoning sequences that can be directly traced in the conceptual diagram. This contrasts with NL-QA, whose reasoning cannot be similarly verified, leaving users and experts uncertain of how and why specific processes were mentioned or omitted.

Taken together, these findings demonstrate that integrating conceptual modeling into the QA framework moves beyond improving standard performance metrics. The introduction of quantitative transparency metrics, supported by direct comparisons in both textual and visual forms, underscores how OPM-QA’s answers are not just better in terms of correctness, but also clearer, more verifiable, and more trustworthy. This alignment of reasoning with a conceptual backbone is particularly valuable in complex domains where understanding the logic behind an answer is as important as the answer itself. As a result, the synergy between neuro-symbolic reasoning and OPM-based conceptual structures offers a promising avenue toward AI systems that users and domain experts can scrutinize, trust, and ultimately shape with confidence.

\section{Conclusion}
\label{sec:conclusion}

We propose Neuro-Conceptual Artificial Intelligence (NCAI), a neuro-symbolic approach that integrates OPM conceptual modeling with deep learning to overcome limitations in traditional knowledge representation and reasoning. By embedding OPM-based conceptual logic into a QA system, NCAI captures complex processes and state changes that conventional triplet-based representations and black box neural models struggle to address. Through this structured, bimodal OPM representation, NCAI provides not only improved answer accuracy but also a demonstrably transparent and interpretable reasoning pathway. The introduction of transparency metrics ($\text{P}_\text{T}$, $\text{R}_\text{T}$, $\text{F1}_\text{T}$) offers quantitative support for the alignment with OPM-defined conceptual structures, moving beyond purely qualitative assessments of interpretability.

Our experimental results demonstrate that NCAI substantially outperforms traditional methods on both standard accuracy-based measures and transparency-focused metrics. By leveraging OPM as a symbolic backbone and employing the LLM under structured guidance, NCAI brings neuro-symbolic AI closer to genuine explainability. Although this work focuses on QA, our conceptual modeling approach may generalize to other tasks requiring robust and interpretable reasoning. Future research will examine scalability to larger, more complex domains, refine prompt designs to handle richer conceptual structures, and integrate our approach with emerging prompting and agentic frameworks. The code and dataset are available on \url{https://github.com/kangxin/NCAI}.

\section*{Limitations}

One limitation of our study is that it relies on a relatively small, self-constructed dataset of 50 question-answer pairs. While sufficient for an initial proof of concept, the generalizability and scalability of NCAI to larger and more complex real-world scenarios remain to be explored. In future work, we intend to evaluate NCAI on larger publicly available benchmarks and more intricate conceptual domains, potentially requiring more efficient prompt designs or incremental model updates to handle extensive OPM knowledge.

Additionally, although QA serves as a proof-of-concept task to demonstrate the feasibility of integrating OPM with LLM, applying this approach to other downstream tasks, such as predictive modeling and real-time decision-making in dynamic environments, would require additional domain-specific adaptations and possibly integration with external data sources. While the OPM-based reasoning structure holds promise beyond QA, confirming its utility in these broader contexts remains an area for future investigation.

Moreover, while our method improves transparency through OPM-driven conceptual alignment, certain ambiguities in the source text can still challenge the strict adherence of the LLM to OPM syntax and conventions. The generated OPM representations might require subsequent refinement by human modelers or more specialized training to ensure full syntactic correctness. Developing standardized benchmarks and further metrics for reasoning transparency, as well as exploring more advanced prompting and agentic design patterns, can help refine the approach, but these steps also remain as future endeavors.

\section*{Acknowledgments}

This research has been supported by the Project of Discretionary Budget of the Dean, Graduate School of Technology, Industrial and Social Sciences, Tokushima University.

\bibliography{custom}

\begin{thebibliography}{37}
\providecommand{\natexlab}[1]{#1}

\bibitem[{Besold et~al.(2017)Besold, Garcez et~al.}]{Besold2017NeuralSymbolic}
Thomas~R Besold, Artur S~d'Avila Garcez, et~al. 2017.
\newblock Neural-symbolic learning and reasoning: A survey and interpretation.
\newblock \emph{arXiv preprint arXiv:1711.03902}.

\bibitem[{Brown et~al.(2023)Brown, Godfrey, Konz, Tu, and Kvinge}]{brown-etal-2023-understanding}
Davis Brown, Charles Godfrey, Nicholas Konz, Jonathan Tu, and Henry Kvinge. 2023.
\newblock \href {https://doi.org/10.18653/v1/2023.emnlp-main.403} {Understanding the inner-workings of language models through representation dissimilarity}.
\newblock In \emph{Proceedings of the 2023 Conference on Empirical Methods in Natural Language Processing}, pages 6543--6558, Singapore. Association for Computational Linguistics.

\bibitem[{De~Cao et~al.(2021)De~Cao, Aziz, and Titov}]{de-cao-etal-2021-editing}
Nicola De~Cao, Wilker Aziz, and Ivan Titov. 2021.
\newblock \href {https://doi.org/10.18653/v1/2021.emnlp-main.522} {Editing factual knowledge in language models}.
\newblock In \emph{Proceedings of the 2021 Conference on Empirical Methods in Natural Language Processing}, pages 6491--6506, Online and Punta Cana, Dominican Republic. Association for Computational Linguistics.

\bibitem[{Ding et~al.(2024{\natexlab{a}})Ding, Kang, and Ren}]{ding2024neuro}
Fei Ding, Xin Kang, and Fuji Ren. 2024{\natexlab{a}}.
\newblock \href {https://doi.org/10.1109/TAFFC.2023.3279318} {Neuro or symbolic? fine-tuned transformer with unsupervised lda topic clustering for text sentiment analysis}.
\newblock \emph{IEEE Transactions on Affective Computing}, 15(2):493--507.

\bibitem[{Ding et~al.(2024{\natexlab{b}})Ding, Kang, Wang, Wu, Nakagawa, and Ren}]{ding2024causal}
Fei Ding, Xin Kang, Linhuang Wang, Yunong Wu, Satoshi Nakagawa, and Fuji Ren. 2024{\natexlab{b}}.
\newblock \href {https://doi.org/10.3390/electronics13091746} {Causal inference and prefix prompt engineering based on text generation models for financial argument analysis}.
\newblock \emph{Electronics}, 13(9):1746.

\bibitem[{Dori(2002)}]{dori2002object}
Dov Dori. 2002.
\newblock \emph{Object-Process Methodology: A Holistic Systems Paradigm; with CD-ROM}.
\newblock Springer Science \& Business Media.

\bibitem[{Dori et~al.(2018)Dori, Jbara, Levi, and Wengrowicz}]{dori2018object}
Dov Dori, Ahmad Jbara, Natali Levi, and Niva Wengrowicz. 2018.
\newblock \href {https://www.ppi-int.com/wp-content/uploads/2018/01/SyEN_61.pdf} {Object-process methodology, opm iso 19450--opcloud and the evolution of opm modeling tools}.
\newblock \emph{Systems Engineering Letters, Project Performance International (PPI) SyEN}, 61:6--17.

\bibitem[{Dori and Shteingardt(2025)}]{dori2025converting}
Dov Dori and Veronika Shteingardt. 2025.
\newblock Converting knowledge from text to opm models using generative ai prompts: The neuro-conceptual approach.
\newblock In \emph{Israel Data Science and AI Initiative 4th Annual Conference}. IDSAI.

\bibitem[{Dori et~al.(2016)}]{dori2016model}
Dov Dori et~al. 2016.
\newblock \emph{Model-based systems engineering with OPM and SysML}, volume~15.
\newblock Springer.

\bibitem[{Doshi-Velez and Kim(2017)}]{DoshiVelez2017RigorousScience}
Finale Doshi-Velez and Been Kim. 2017.
\newblock Towards a rigorous science of interpretable machine learning.
\newblock \emph{arXiv preprint arXiv:1702.08608}.

\bibitem[{Dou and Kang(2024)}]{dou2024tam}
Rongyu Dou and Xin Kang. 2024.
\newblock \href {https://doi.org/10.1016/j.compeleceng.2023.109071} {{TAM-SenticNet}: A neuro-symbolic ai approach for early depression detection via social media analysis}.
\newblock \emph{Computers and Electrical Engineering}, 114:109071.

\bibitem[{El~Zini and Awad(2022)}]{el-zini-awad-2022-beyond}
Julia El~Zini and Mariette Awad. 2022.
\newblock \href {https://doi.org/10.18653/v1/2022.findings-emnlp.100} {Beyond model interpretability: On the faithfulness and adversarial robustness of contrastive textual explanations}.
\newblock In \emph{Findings of the Association for Computational Linguistics: EMNLP 2022}, pages 1391--1402, Abu Dhabi, United Arab Emirates. Association for Computational Linguistics.

\bibitem[{Feng et~al.(2024)Feng, Zhou, Zhu, and Mao}]{feng-etal-2024-promptexplainer}
Zijian Feng, Hanzhang Zhou, Zixiao Zhu, and Kezhi Mao. 2024.
\newblock \href {https://aclanthology.org/2024.findings-eacl.60} {{P}rompt{E}xplainer: Explaining language models through prompt-based learning}.
\newblock In \emph{Findings of the Association for Computational Linguistics: EACL 2024}, pages 882--895, St. Julian's, Malta. Association for Computational Linguistics.

\bibitem[{Garcez and Lamb(2023)}]{Garcez2023NeuroSymbolic}
Artur~d’Avila Garcez and Luis~C Lamb. 2023.
\newblock Neurosymbolic ai: The 3 rd wave.
\newblock \emph{Artificial Intelligence Review}, 56(11):12387--12406.

\bibitem[{Heinzerling and Inui(2021)}]{heinzerling-inui-2021-language}
Benjamin Heinzerling and Kentaro Inui. 2021.
\newblock \href {https://doi.org/10.18653/v1/2021.eacl-main.153} {Language models as knowledge bases: On entity representations, storage capacity, and paraphrased queries}.
\newblock In \emph{Proceedings of the 16th Conference of the European Chapter of the Association for Computational Linguistics: Main Volume}, pages 1772--1791, Online. Association for Computational Linguistics.

\bibitem[{Horovicz and Goldshmidt(2024)}]{horovicz-goldshmidt-2024-tokenshap}
Miriam Horovicz and Roni Goldshmidt. 2024.
\newblock \href {https://doi.org/10.18653/v1/2024.nlp4science-1.1} {{T}oken{SHAP}: Interpreting large language models with {M}onte {C}arlo shapley value estimation}.
\newblock In \emph{Proceedings of the 1st Workshop on NLP for Science (NLP4Science)}, pages 1--8, Miami, FL, USA. Association for Computational Linguistics.

\bibitem[{Kanaani et~al.(2024)Kanaani, Dadkhah, and Ghorbani}]{kanaani-etal-2024-triple}
Mohammadamin Kanaani, Sajjad Dadkhah, and Ali~A. Ghorbani. 2024.
\newblock \href {https://aclanthology.org/2024.lrec-main.1463} {Triple-{R}: Automatic reasoning for fact verification using language models}.
\newblock In \emph{Proceedings of the 2024 Joint International Conference on Computational Linguistics, Language Resources and Evaluation (LREC-COLING 2024)}, pages 16831--16840, Torino, Italia. ELRA and ICCL.

\bibitem[{Kang(2024)}]{kang2024transparency}
Xin Kang. 2024.
\newblock {Transparency in the dimABSA Task with Neuro-Symbolic and Generative AI}.
\newblock \emph{IEICE Technical Report}, 124(90(CQ2024 16-41)):74--79.

\bibitem[{Kang et~al.(2024)Kang, Zhang, Zhou, Wu, Shi, and Matsumoto}]{kang-etal-2024-tmak}
Xin Kang, Zhifei Zhang, Jiazheng Zhou, Yunong Wu, Xuefeng Shi, and Kazuyuki Matsumoto. 2024.
\newblock {TMAK}-plus at {SIGHAN}-2024 dim{ABSA} task: Multi-agent collaboration for transparent and rational sentiment analysis.
\newblock In \emph{Proceedings of the 10th SIGHAN Workshop on Chinese Language Processing (SIGHAN-10)}, pages 88--95, Bangkok, Thailand. Association for Computational Linguistics.

\bibitem[{Kohen and Dori(2021)}]{kohen2021designing}
Hanan Kohen and Dov Dori. 2021.
\newblock \href {https://doi.org/10.20935/AL1918} {Designing and developing opcloud, an opm-based collaborative software environment, in a mixed academic and industrial setting: An experience report}.
\newblock \emph{Academia Letters}.

\bibitem[{Lin(2004)}]{lin-2004-rouge}
Chin-Yew Lin. 2004.
\newblock {ROUGE}: A package for automatic evaluation of summaries.
\newblock In \emph{Text Summarization Branches Out}, pages 74--81, Barcelona, Spain. Association for Computational Linguistics.

\bibitem[{Lipton(2018)}]{Lipton2018Mythos}
Zachary~C Lipton. 2018.
\newblock The mythos of model interpretability: In machine learning, the concept of interpretability is both important and slippery.
\newblock \emph{Queue}, 16(3):31--57.

\bibitem[{Mosca et~al.(2022)Mosca, Szigeti, Tragianni, Gallagher, and Groh}]{mosca-etal-2022-shap}
Edoardo Mosca, Ferenc Szigeti, Stella Tragianni, Daniel Gallagher, and Georg Groh. 2022.
\newblock \href {https://aclanthology.org/2022.coling-1.406} {{SHAP}-based explanation methods: A review for {NLP} interpretability}.
\newblock In \emph{Proceedings of the 29th International Conference on Computational Linguistics}, pages 4593--4603, Gyeongju, Republic of Korea. International Committee on Computational Linguistics.

\bibitem[{Oguz et~al.(2022)Oguz, Chen, Karpukhin, Peshterliev, Okhonko, Schlichtkrull, Gupta, Mehdad, and Yih}]{oguz-etal-2022-unik}
Barlas Oguz, Xilun Chen, Vladimir Karpukhin, Stan Peshterliev, Dmytro Okhonko, Michael Schlichtkrull, Sonal Gupta, Yashar Mehdad, and Scott Yih. 2022.
\newblock \href {https://doi.org/10.18653/v1/2022.findings-naacl.115} {{U}ni{K}-{QA}: Unified representations of structured and unstructured knowledge for open-domain question answering}.
\newblock In \emph{Findings of the Association for Computational Linguistics: NAACL 2022}, pages 1535--1546, Seattle, United States. Association for Computational Linguistics.

\bibitem[{Rudin(2019)}]{Rudin2019StopExplaining}
Cynthia Rudin. 2019.
\newblock Stop explaining black box machine learning models for high stakes decisions and use interpretable models instead.
\newblock \emph{Nature Machine Intelligence}, 1(5):206--215.

\bibitem[{Sellam et~al.(2020)Sellam, Das, and Parikh}]{sellam-etal-2020-bleurt}
Thibault Sellam, Dipanjan Das, and Ankur Parikh. 2020.
\newblock {BLEURT}: Learning robust metrics for text generation.
\newblock In \emph{Proceedings of the 58th Annual Meeting of the Association for Computational Linguistics}, pages 7881--7892, Online. Association for Computational Linguistics.

\bibitem[{Shi et~al.(2021)Shi, Cao, Hou, Li, and Zhang}]{shi-etal-2021-transfernet}
Jiaxin Shi, Shulin Cao, Lei Hou, Juanzi Li, and Hanwang Zhang. 2021.
\newblock \href {https://doi.org/10.18653/v1/2021.emnlp-main.341} {{T}ransfer{N}et: An effective and transparent framework for multi-hop question answering over relation graph}.
\newblock In \emph{Proceedings of the 2021 Conference on Empirical Methods in Natural Language Processing}, pages 4149--4158, Online and Punta Cana, Dominican Republic. Association for Computational Linguistics.

\bibitem[{Tjoa and Guan(2020)}]{Tjoa2020SurveyXAI}
Eliot Tjoa and Cuntai Guan. 2020.
\newblock A survey on explainable artificial intelligence (xai): Towards medical xai.
\newblock \emph{arXiv preprint arXiv:1907.07374}.

\bibitem[{Wang et~al.(2022)Wang, Shen, Peng, Zhang, Xiao, Liu, Tang, Chen, Wu, and Wang}]{wang-etal-2022-fine}
Lijie Wang, Yaozong Shen, Shuyuan Peng, Shuai Zhang, Xinyan Xiao, Hao Liu, Hongxuan Tang, Ying Chen, Hua Wu, and Haifeng Wang. 2022.
\newblock \href {https://doi.org/10.18653/v1/2022.conll-1.6} {A fine-grained interpretability evaluation benchmark for neural {NLP}}.
\newblock In \emph{Proceedings of the 26th Conference on Computational Natural Language Learning (CoNLL)}, pages 70--84, Abu Dhabi, United Arab Emirates (Hybrid). Association for Computational Linguistics.

\bibitem[{Wang et~al.(2017)Wang, Mao, Wang, and Guo}]{Wang2017KnowledgeGraphEmbedding}
Quan Wang, Zhendong Mao, Bin Wang, and Li~Guo. 2017.
\newblock Knowledge graph embedding: A survey of approaches and applications.
\newblock \emph{IEEE Transactions on Knowledge and Data Engineering}, 29(12):2724--2743.

\bibitem[{West et~al.(2022)West, Bhagavatula, Hessel, Hwang, Jiang, Le~Bras, Lu, Welleck, and Choi}]{west-etal-2022-symbolic}
Peter West, Chandra Bhagavatula, Jack Hessel, Jena Hwang, Liwei Jiang, Ronan Le~Bras, Ximing Lu, Sean Welleck, and Yejin Choi. 2022.
\newblock \href {https://doi.org/10.18653/v1/2022.naacl-main.341} {Symbolic knowledge distillation: from general language models to commonsense models}.
\newblock In \emph{Proceedings of the 2022 Conference of the North American Chapter of the Association for Computational Linguistics: Human Language Technologies}, pages 4602--4625, Seattle, United States. Association for Computational Linguistics.

\bibitem[{Yang et~al.(2022)Yang, Zhang, Erfani, and Lau}]{yang-etal-2022-interpretable}
Shiquan Yang, Rui Zhang, Sarah Erfani, and Jey~Han Lau. 2022.
\newblock \href {https://doi.org/10.18653/v1/2022.acl-long.338} {An interpretable neuro-symbolic reasoning framework for task-oriented dialogue generation}.
\newblock In \emph{Proceedings of the 60th Annual Meeting of the Association for Computational Linguistics (Volume 1: Long Papers)}, pages 4918--4935, Dublin, Ireland. Association for Computational Linguistics.

\bibitem[{Yasunaga et~al.(2021)Yasunaga, Ren, Bosselut, Liang, and Leskovec}]{yasunaga-etal-2021-qa}
Michihiro Yasunaga, Hongyu Ren, Antoine Bosselut, Percy Liang, and Jure Leskovec. 2021.
\newblock \href {https://doi.org/10.18653/v1/2021.naacl-main.45} {{QA}-{GNN}: Reasoning with language models and knowledge graphs for question answering}.
\newblock In \emph{Proceedings of the 2021 Conference of the North American Chapter of the Association for Computational Linguistics: Human Language Technologies}, pages 535--546, Online. Association for Computational Linguistics.

\bibitem[{Zhang et~al.(2023)Zhang, Huo, Elazar, Song, Goldberg, and Roth}]{zhang-etal-2023-cikqa}
Hongming Zhang, Yintong Huo, Yanai Elazar, Yangqiu Song, Yoav Goldberg, and Dan Roth. 2023.
\newblock \href {https://doi.org/10.18653/v1/2023.findings-eacl.8} {{CIKQA}: Learning commonsense inference with a unified knowledge-in-the-loop {QA} paradigm}.
\newblock In \emph{Findings of the Association for Computational Linguistics: EACL 2023}, pages 114--124, Dubrovnik, Croatia. Association for Computational Linguistics.

\bibitem[{Zhang et~al.(2022)Zhang, Dai, Dong, and He}]{zhang-etal-2022-drlk}
Miao Zhang, Rufeng Dai, Ming Dong, and Tingting He. 2022.
\newblock \href {https://doi.org/10.18653/v1/2022.emnlp-main.342} {{DRLK}: Dynamic hierarchical reasoning with language model and knowledge graph for question answering}.
\newblock In \emph{Proceedings of the 2022 Conference on Empirical Methods in Natural Language Processing}, pages 5123--5133, Abu Dhabi, United Arab Emirates. Association for Computational Linguistics.

\bibitem[{Zhao et~al.(2024)Zhao, Chen, Cohan, and Zhao}]{zhao-etal-2024-tapera}
Yilun Zhao, Lyuhao Chen, Arman Cohan, and Chen Zhao. 2024.
\newblock \href {https://doi.org/10.18653/v1/2024.acl-long.692} {{T}a{PERA}: Enhancing faithfulness and interpretability in long-form table {QA} by content planning and execution-based reasoning}.
\newblock In \emph{Proceedings of the 62nd Annual Meeting of the Association for Computational Linguistics (Volume 1: Long Papers)}, pages 12824--12840, Bangkok, Thailand. Association for Computational Linguistics.

\bibitem[{Zhu et~al.(2024)Zhu, Hwang, Dugan, and Callison-Burch}]{zhu-etal-2024-fanoutqa}
Andrew Zhu, Alyssa Hwang, Liam Dugan, and Chris Callison-Burch. 2024.
\newblock \href {https://doi.org/10.18653/v1/2024.acl-short.2} {{F}an{O}ut{QA}: A multi-hop, multi-document question answering benchmark for large language models}.
\newblock In \emph{Proceedings of the 62nd Annual Meeting of the Association for Computational Linguistics (Volume 2: Short Papers)}, pages 18--37, Bangkok, Thailand. Association for Computational Linguistics.

\end{thebibliography}

\clearpage

\onecolumn

\appendix

\section{Natural Language Text}
\label{sec:nlt}

\begin{tcolorbox}[breakable, title=Natural Language Text, colframe=black, colback=white, label=box:nl_text]
A plausible set of concise steps of how the process that transforms informal rules of thumb into well-established principles that guide systems engineering practice follows.
\begin{enumerate}[label=\arabic*., leftmargin=*, nosep, itemsep=0pt, topsep=0pt]
    \item Initial observation: Heuristics often start as informal rules of thumb based on practical experience.
    \item Documentation and sharing: These observations get documented and shared among practitioners.
    \item Testing and refinement: The heuristics are tested in various projects and refined based on outcomes.
    \item Pattern recognition: As similar heuristics prove useful across multiple projects and domains, recognizable patterns emerge, enabling heuristic generalization.
    \item Formal studies: Researchers conduct formal studies to validate the effectiveness of the heuristic.
    \item Theoretical backing: The heuristics are connected to underlying theories in systems engineering and related fields.
    \item Consensus building: As evidence accumulates, a consensus forms in the systems engineering community about the validity and importance of the heuristic.
\end{enumerate}
\end{tcolorbox}

\section{OPL of Constructed OPM Model}
\label{sec:ocom}

\begin{tcolorbox}[breakable, title=OPL for System Diagram (SD), colframe=black, colback=white, label=box:manual_opl_sd]
\begin{enumerate}[label=\arabic*., leftmargin=*, nosep, itemsep=0pt, topsep=0pt]
    \item Heuristic can be principle, rule of thumb or at one of five other states. State rule of thumb is initial. State principle is final.
    \item Heuristic-to-principle Evolving changes Heuristic from rule of thumb to principle.
    \item Systems Engineering Practitioner \& Expert Group handles Heuristic-to-principle Evolving.
\end{enumerate}
\end{tcolorbox}

\begin{tcolorbox}[breakable, title=OPL for In-Zoomed Diagram (SD1), colframe=black, colback=white, label=box:manual_opl_sd1]
\begin{enumerate}[label=\arabic*., widest=99, leftmargin=*, nosep, itemsep=0pt, topsep=0pt]
    \item Heuristic-to-principle Evolving from SD zooms in SD1 into Initial Observing, Documenting \& Sharing, Project Selecting, Testing \& Refining, Pattern Emerging \& Recognizing, Effectiveness Validating, Theoretical Baking, and Consensus Building, which occur in that time sequence.
    \item Heuristic can be documented \& shared, effectiveness validated, pattern recognized, principle, rule of thumb, tested \& refined or theoretically backed. State rule of thumb is initial. State principle is final.
    \item Systems Engineering Practitioner \& Expert Group handles Heuristic-to-principle Evolving.
    \item Documenting \& Sharing changes Heuristic from rule of thumb to documented \& shared.
    \item Testing \& Refining changes Heuristic from documented \& shared to tested \& refined.
    \item Testing \& Refining requires Project Set.
    \item Pattern Emerging \& Recognizing changes Heuristic from tested \& refined to pattern recognized.
    \item Pattern Emerging \& Recognizing requires Project Set.
    \item Effectiveness Validating changes Heuristic from pattern recognized to effectiveness validated.
    \item Effectiveness Validating requires Project Set.
    \item Theoretical Baking changes Heuristic from effectiveness validated to theoretically backed.
    \item Consensus Building changes Heuristic from theoretically backed to principle.
    \item Initial Observing changes Heuristic to state rule of thumb.
    \item Project Selecting yields Project Set.
\end{enumerate}
\end{tcolorbox}

\section{OPL Generated by LLM}
\label{sec:ogl}

\begin{tcolorbox}[breakable, title=OPL Generated by LLM, colframe=black, colback=white, label=box:llm_opl]
\begin{enumerate}[label=\arabic*., widest=99, leftmargin=*, nosep, itemsep=0pt, topsep=0pt]
    \item Principle Establishing changes Heuristic from rule of thumb to principle.
    \item Practitioner handles Documenting.
    \item Practitioner handles Sharing.
    \item Researcher handles Formal Studying.
    \item Systems Engineering Community handles Consensus Building.
    \item Principle Establishing zooms into Observing, Documenting, Sharing, Testing, Refining, Pattern Recognizing, Formal Studying, Connecting, and Consensus Building, which occur in that time sequence.
    \item Heuristic can be rule of thumb, documented, shared, tested, refined, recognized pattern, validated, theorized, principle.
    \item The state rule of thumb is initial. State principle is final.
    \item Observing yields Heuristic.
    \item Documenting changes Heuristic from rule of thumb to documented.
    \item Sharing changes Heuristic from documented to shared.
    \item Testing changes Heuristic from shared to tested.
    \item Refining changes Heuristic from tested to refined.
    \item Testing requires Project.
    \item Refining consumes Outcome.
    \item Pattern Recognizing changes Heuristic from refined to recognized pattern.
    \item Formal Studying changes Heuristic from recognized pattern to validated.
    \item Connecting changes Heuristic from validated to theorized.
    \item Connecting consumes Theory.
    \item Consensus Building changes Heuristic from theorized to principle.
\end{enumerate}
\end{tcolorbox}

\section{Prompt for the QA System}
\label{sec:qa_prompt}

\begin{tcolorbox}[breakable, title=Prompt for OPM-QA and NL-QA Systems, colframe=black, colback=white, label=box:prompt_qa]
\noindent
\textbf{Prompt:}\\[6pt]
You are a knowledgeable assistant. 
Your task is to answer questions based on the provided domain knowledge. 
Your answers should align closely with the domain knowledge, use precise terminology, and remain concise and accurate.
Focus on identifying and describing key processes, objects, and states explicitly, and clarify their relationships where relevant.\\[6pt]
\textbf{Domain Knowledge:}\\[6pt]
[OPL Knowledge in Appendix~\ref{sec:ocom} or NL Knowledge in Appendix~\ref{sec:nlt}]\\[6pt]
\textbf{Examples of Question-Answer Pairs:}\\[6pt]
Q: [example question 1]\\
A: [example answer 1]\\
...\\
Q: [example question N]\\
A: [example answer N]\\[6pt]
\textbf{New Question:}\\[6pt]
Q: [question]\\
A (concise and precise):
\end{tcolorbox}

\clearpage

\begin{landscape}

\section{Examples of Questions, Answers, and Evaluation Results}
\label{appendix:qa_results}

\begin{table}[H]
\centering 
\caption{10 example questions and ground truth answers from the QA dataset.}
\label{tab:qa_table}

\setlength{\tabcolsep}{4pt}    
\renewcommand{\arraystretch}{1.4} 

\begin{tabular}{|c|p{9cm}|p{14cm}|}
\hline
\textbf{ID} & \textbf{Question} & \textbf{Ground Truth Answer} \\
\hline
1 & What is the relationship between Testing \& Refining and Pattern Emerging \& Recognizing in Heuristic evolution? & Testing \& Refining changes Heuristic from documented \& shared to tested \& refined, and Pattern Emerging \& Recognizing then changes it from tested \& refined to pattern recognized. \\ \hline
2 & How does Heuristic achieve theoretical backing before becoming a principle? & Heuristic achieves theoretical backing by undergoing Formal Studying, which changes it from pattern recognized to effectiveness validated, followed by Theoretical Baking, which changes it to theoretically backed, and finally Consensus Building to become a principle. \\ \hline
3 & How does Heuristic change from effectiveness validated to principle? & Heuristic changes from effectiveness validated to principle through Theoretical Baking and Consensus Building. \\ \hline
4 & How does the Heuristic-to-priniciple Evolving process relate to the different states of Heuristic? & The Heuristic-to-priniciple Evolving process changes Heuristic from rule of thumb through documented \& shared, tested \& refined, pattern recognized, effectiveness validated, theoretically backed, and finally to principle. \\ \hline
5 & What processes change Heuristic from rule of thumb to pattern recognized? & Heuristic changes from rule of thumb to pattern recognized through Documenting \& Sharing, Testing \& Refining, and Pattern Emerging \& Recognizing processes. \\ \hline
6 & What processes change Heuristic from rule of thumb to tested \& refined? & Heuristic changes from rule of thumb to tested \& refined through Documenting \& Sharing and Testing \& Refining processes. \\ \hline
7 & How does Heuristic transition from effectiveness validated to principle? & Heuristic transitions from effectiveness validated to principle through Theoretical Baking and Consensus Building. \\ \hline
8 & How does Heuristic transition from documented \& shared to pattern recognized? & Heuristic transitions from documented \& shared to pattern recognized through Testing \& Refining and Pattern Emerging \& Recognizing processes. \\ \hline
9 & How does Heuristic evolve from documented \& shared to effectiveness validated? & Heuristic evolves from documented \& shared to effectiveness validated through Testing \& Refining, Pattern Emerging \& Recognizing, and Formal Studying processes. \\ \hline
10 & What processes are involved in Heuristic evolving from rule of thumb to effectiveness validated? & Heuristic evolves from rule of thumb to effectiveness validated through Documenting \& Sharing, Testing \& Refining, Pattern Emerging \& Recognizing, and Formal Studying processes. \\ \hline
\end{tabular}
\end{table}

\end{landscape}

\clearpage

\begin{landscape}

\begin{table}[p]
\centering 
\caption{Evaluation results of OPM-QA for 10 examples. Performance was measured using Loose Accuracy (LA), Strict Accuracy (SA), ROUGE-1 (R-1), ROUGE-2 (R-2), ROUGE-L (R-L), BLEURT (BT), the GPT Judge Score (GPT),  Transparency Precision ($\text{P}_\text{T}$) , Transparency Recall ($\text{R}_\text{T}$), and Transparency F1 ($\text{F1}_\text{T}$).}
\label{tab:results_opl}

\setlength{\tabcolsep}{2pt}    
\renewcommand{\arraystretch}{1.1} 

\begin{tabular}{|c|P{13.9cm}|c|c|c|c|c|c|c|c|c|c|}
\hline
\rule{0pt}{2.5ex} 
\textbf{ID} & \textbf{Answer} & \textbf{LA} & \textbf{SA} & \textbf{R-1} & \textbf{R-2} & \textbf{R-L} & \textbf{BT} & \textbf{GPT} & $\textbf{P}_\textbf{T}$ & $\textbf{R}_\textbf{T}$ & $\textbf{F1}_\textbf{T}$ \\
\hline
1 & Testing \& Refining precedes Pattern Emerging \& Recognizing in Heuristic evolution. Testing \& Refining changes Heuristic from documented \& shared to tested \& refined, while Pattern Emerging \& Recognizing subsequently changes Heuristic from tested \& refined to pattern recognized. Both processes require a Project Set. & 1.000 & 1.000 & 0.696 & 0.591 & 0.696 & 0.444 & 1.000 & 1.000 & 1.000 & 1.000 \\ \hline
2 & Heuristic achieves theoretical backing through Theoretical Baking, which changes it from effectiveness validated to theoretically backed. Subsequently, it undergoes Consensus Building to transition from theoretically backed to principle. & 0.619 & 0.487 & 0.714 & 0.400 & 0.571 & 0.496 & 1.000 & 1.000 & 1.000 & 1.000 \\ \hline
3 & Heuristic changes from effectiveness validated to principle through Theoretical Baking, which transitions it to theoretically backed, followed by Consensus Building, which transitions it to principle. & 1.000 & 1.000 & 0.750 & 0.636 & 0.750 & 0.461 & 1.000 & 0.800 & 1.000 & 0.889 \\ \hline
4 & The Heuristic-to-principle Evolving process transitions Heuristic through the states of rule of thumb (initial), documented \& shared, tested \& refined, pattern recognized, effectiveness validated, theoretically backed, and finally to principle (final), following a defined sequence of processes. & 0.895 & 0.846 & 0.756 & 0.605 & 0.756 & 0.583 & 1.000 & 1.000 & 0.857 & 0.923 \\ \hline
5 & The processes that change Heuristic from rule of thumb to pattern recognized are Documenting \& Sharing, Testing \& Refining, and Pattern Emerging \& Recognizing. & 1.000 & 1.000 & 1.000 & 0.769 & 0.857 & 0.808 & 1.000 & 1.000 & 1.000 & 1.000 \\ \hline
6 & The processes that change Heuristic from rule of thumb to tested \& refined are Documenting \& Sharing followed by Testing \& Refining. & 1.000 & 1.000 & 0.957 & 0.571 & 0.783 & 0.753 & 1.000 & 1.000 & 1.000 & 1.000 \\ \hline
7 & Heuristic transitions from effectiveness validated to principle through Theoretical Baking, which changes it to theoretically backed, followed by Consensus Building, which changes it to principle. & 1.000 & 1.000 & 0.750 & 0.636 & 0.750 & 0.463 & 1.000 & 0.800 & 1.000 & 0.889 \\ \hline
8 & Heuristic transitions from documented \& shared to pattern recognized through the processes of Testing \& Refining, which changes it to tested \& refined, followed by Pattern Emerging \& Recognizing, which changes it to pattern recognized. & 1.000 & 1.000 & 0.774 & 0.621 & 0.710 & 0.537 & 1.000 & 0.667 & 1.000 & 0.800 \\ \hline
9 & Heuristic evolves from documented \& shared to effectiveness validated through the processes of Testing \& Refining, Pattern Emerging \& Recognizing, and Effectiveness Validating, in that sequence. & 0.857 & 0.794 & 0.828 & 0.667 & 0.759 & 0.634 & 1.000 & 0.750 & 1.000 & 0.857 \\ \hline
10 & The processes involved in Heuristic evolving from rule of thumb to effectiveness validated are Initial Observing, Documenting \& Sharing, Testing \& Refining, Pattern Emerging \& Recognizing, and Effectiveness Validating. & 0.812 & 0.732 & 0.800 & 0.667 & 0.743 & 0.500 & 1.000 & 0.667 & 1.000 & 0.800 \\ \hline
\end{tabular}
\end{table}

\end{landscape}

\clearpage

\begin{landscape}

\begin{table}[p]
\centering 
\caption{Evaluation results of NL-QA for 10 examples. Performance was measured using Loose Accuracy (LA), Strict Accuracy (SA), ROUGE-1 (R‑1), ROUGE-2 (R‑2), ROUGE-L (R‑L), BLEURT (BT), the GPT Judge Score (GPT), Transparency Precision ($\text{P}_\text{T}$), Transparency Recall ($\text{R}_\text{T}$), and Transparency F1 ($\text{F1}_\text{T}$)}
\label{tab:results_nl}

\setlength{\tabcolsep}{2pt}    
\renewcommand{\arraystretch}{1.1} 

\begin{tabular}{|c|P{13.9cm}|c|c|c|c|c|c|c|c|c|c|}
\hline
\rule{0pt}{2.5ex} 
\textbf{ID} & \textbf{Answer} & \textbf{LA} & \textbf{SA} & \textbf{R-1} & \textbf{R-2} & \textbf{R-L} & \textbf{BT} & \textbf{GPT} & $\textbf{P}_\textbf{T}$ & $\textbf{R}_\textbf{T}$ & $\textbf{F1}_\textbf{T}$ \\
\hline
1 & Testing \& Refining provides the practical outcomes and data that enable Pattern Emerging \& Recognizing, where recurring success across projects reveals generalizable patterns. & 0.417 & 0.269 & 0.364 & 0.194 & 0.364 & 0.370 & 0.000 & 1.000 & 0.333 & 0.500 \\ \hline
2 & Heuristic achieves theoretical backing through formal studies that validate its effectiveness and by connecting it to underlying theories in systems engineering and related fields. & 0.286 & 0.153 & 0.421 & 0.222 & 0.368 & 0.428 & 1.000 & 0.000 & 0.000 & 0.000 \\ \hline
3 & Heuristic changes from effectiveness validated to principle through Consensus Building, where accumulated evidence and community agreement establish its validity and importance as a guiding principle. & 0.778 & 0.686 & 0.560 & 0.435 & 0.560 & 0.424 & 1.000 & 1.000 & 0.750 & 0.857 \\ \hline
4 & The Heuristic-to-principle Evolving process transitions the heuristic through distinct states: starting as a rule of thumb (Initial Observation), becoming documented and shared (Documentation and Sharing), then tested and refined (Testing and Refinement). It progresses to a pattern recognized state (Pattern Recognition), followed by effectiveness validated (Formal Studies), and finally becomes theoretically backed (Theoretical Backing) and widely accepted through consensus (Consensus Building). Each process step transforms the heuristic into a more formalized and reliable principle. & 0.895 & 0.846 & 0.493 & 0.239 & 0.464 & 0.404 & 1.000 & 0.800 & 0.571 & 0.667 \\ \hline
5 & The processes that change Heuristic from rule of thumb to pattern recognized are Initial Observation, Documentation and Sharing, Testing and Refinement, and Pattern Recognition. & 0.692 & 0.576 & 0.828 & 0.519 & 0.690 & 0.584 & 1.000 & 1.000 & 0.333 & 0.500 \\ \hline
6 & The processes Initial Observation, Documentation and Sharing, and Testing and Refinement change Heuristic from rule of thumb to tested \& refined. & 0.818 & 0.740 & 0.917 & 0.545 & 0.417 & 0.576 & 1.000 & 1.000 & 0.500 & 0.667 \\ \hline
7 & Heuristic transitions from effectiveness validated to principle through Theoretical Backing and Consensus Building, where it is connected to underlying theories and gains community agreement on its validity and importance. & 0.889 & 0.838 & 0.615 & 0.500 & 0.615 & 0.448 & 1.000 & 1.000 & 0.750 & 0.857 \\ \hline
8 & Heuristic transitions from documented \& shared to pattern recognized through Testing \& Refinement, where it is applied across projects, and Pattern Recognition, where recurring usefulness across domains leads to generalization. & 0.636 & 0.508 & 0.581 & 0.483 & 0.581 & 0.509 & 1.000 & 1.000 & 0.500 & 0.667 \\ \hline
9 & Heuristic evolves from documented \& shared to effectiveness validated through Testing \& Refinement, followed by Pattern Recognition and Formal Studies. & 0.643 & 0.515 & 0.815 & 0.640 & 0.815 & 0.705 & 1.000 & 1.000 & 0.667 & 0.800 \\ \hline
10 & The processes involved in Heuristic evolving from rule of thumb to effectiveness validated are Initial Observation, Documentation and Sharing, Testing and Refinement, Pattern Recognition, and Formal Studies. & 0.625 & 0.494 & 0.824 & 0.625 & 0.765 & 0.589 & 1.000 & 1.000 & 0.500 & 0.667 \\ \hline
\end{tabular}
\end{table}

\end{landscape}

\end{document}